\documentclass{article}

\usepackage{eva_report,times}

\usepackage{amsmath}
\usepackage{amssymb}
\usepackage{graphicx}
\usepackage{capt-of}
\usepackage{booktabs}
\usepackage{multirow}
\usepackage{array}
\usepackage{makecell}
\usepackage{colortbl}
\usepackage{xcolor}
\usepackage{wrapfig}
\usepackage{float} %
\usepackage{tabularx}
\usepackage{url}
\usepackage[
    colorlinks=true,
    linkcolor=magenta,
    urlcolor=magenta,
    citecolor=blue
]{hyperref}
\usepackage[capitalize,noabbrev]{cleveref}

\providecommand{\keywords}[1]{}

\title{Aligning Video World Models with Executable Robot Actions via Inverse Dynamics Rewards}

\author{
{\small
\textbf{Ruixiang Wang}\textsuperscript{1,2},\hspace{0.75em}
\textbf{Qingming Liu}\textsuperscript{1},\hspace{0.75em}
\textbf{Yueci Deng}\textsuperscript{1,2},\hspace{0.75em}
\textbf{Guiliang Liu}\textsuperscript{1},\hspace{0.75em}
\textbf{Zhen Liu}\textsuperscript{1}\thanks{Corresponding author.},\hspace{0.75em}
\textbf{Kui Jia}\textsuperscript{1}}\\[0.4em]
\parbox[t]{\textwidth}{\raggedright
\textsuperscript{1}The Chinese University of Hong Kong, Shenzhen\quad
\textsuperscript{2}DexForce}
}

\hypersetup{
    pdftitle={Aligning Video World Models with Executable Robot Actions via Inverse Dynamics Rewards},
    pdfauthor={Ruixiang Wang, Qingming Liu, Yueci Deng, Guiliang Liu, Zhen Liu, Kui Jia},
    pdfsubject={Technical Report}
}

\begin{document}

\maketitle
\suppressfloats[t] %

\begin{abstract}

Video generative models offer a promising basis for robotic planning by predicting future visual rollouts from observations and task instructions. However, their training objectives typically lack explicit constraints on robot executability. Generated rollouts can exhibit embodiment-related artifacts, including arm deformation and temporal discontinuities, while the actions decoded from these rollouts may be unstable or violate robot motion limits. We term this mismatch between generated robot behavior and embodiment-specific execution requirements the \emph{executability gap}. We introduce \textbf{Executable Video Alignment (EVA)}, a post-training framework that uses action-space feedback to improve both the visual integrity of generated robot motion and the feasibility of decoded actions. EVA trains an inverse dynamics model (IDM) on robot demonstrations and freezes it to decode actions from generated videos. We compute a reward from these actions that penalizes non-smooth motion and violations of velocity and acceleration limits, and use it to guide reinforcement-learning updates to the video generator. By feeding decoded-action constraints back into video generation, EVA links visual and action-level improvements through a shared alignment signal. Experiments on RoboTwin, LIBERO, and a real bimanual robot show that EVA reduces embodiment-related visual artifacts and improves task success when executing IDM-decoded actions. 
Further experiments show that video models fine-tuned with EVA generate more effective video--action pairs for training downstream VLAs. Project page: \url{http://eva-project-page.github.io/}

\end{abstract}

\begin{figure}[t]
    \centering
    \includegraphics[width=1\linewidth]{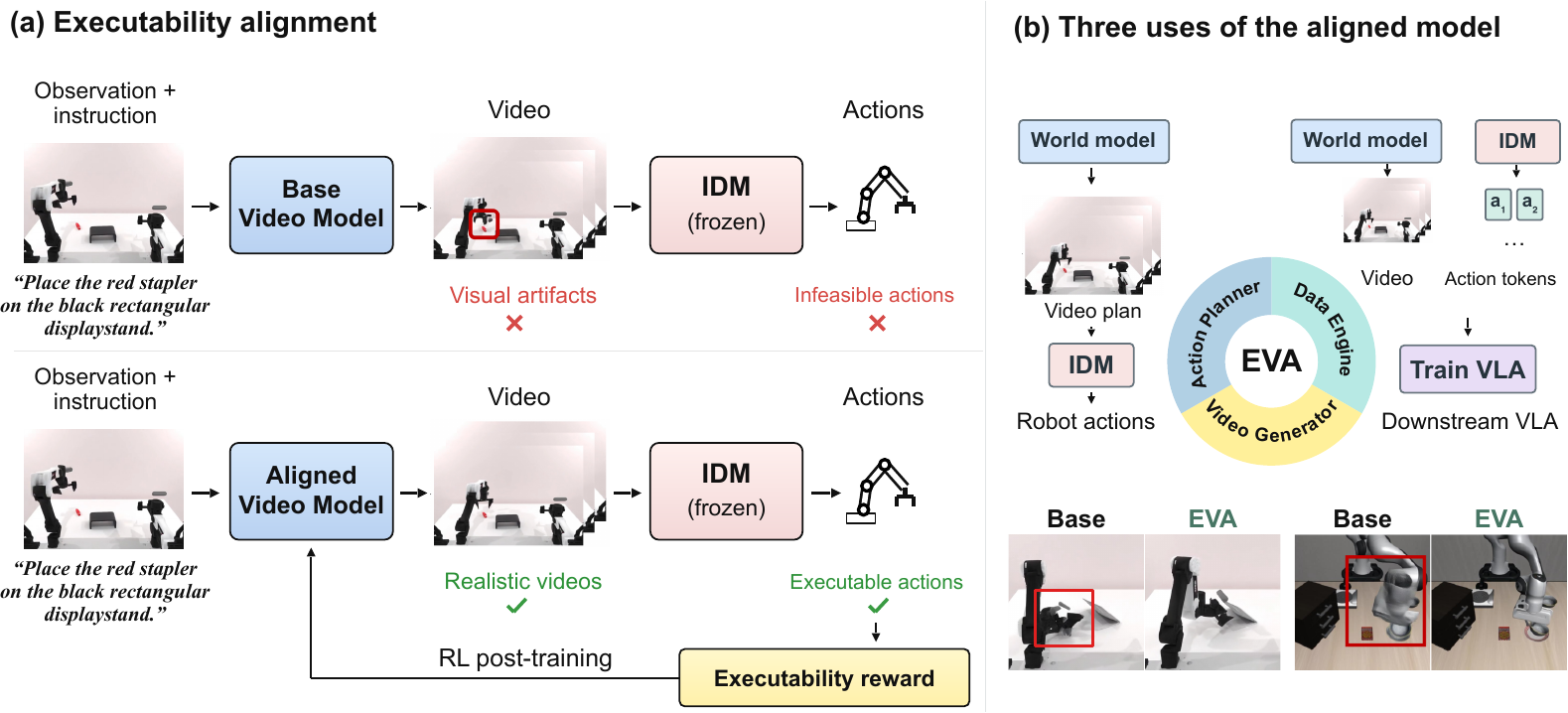}
    \caption{Overview of Executable Video Alignment (EVA).
    (a) EVA post-trains a video generator using rewards computed from actions decoded by a frozen inverse dynamics model (IDM), encouraging smoother, embodiment-consistent motion.
    (b) The aligned model supports video generation and video-based action planning, and can supply video--action pairs for downstream VLA training. The examples compare generated videos before and after alignment.}
    \label{fig:main_figure}
\end{figure}

\section{Introduction}
\label{sec:intro}

Developing generalist robot policies capable of executing diverse manipulation tasks remains a central pursuit in embodied AI. Vision-Language-Action (VLA) models~\citep{black2024pi_0, kim2024openvlaopensourcevisionlanguageactionmodel, brohan2023rt2visionlanguageactionmodelstransfer, gr00tn1_2025} have made significant progress by mapping 2D visual observations and language instructions directly to low-level motor commands. However, scaling robust long-horizon behavior remains challenging when physical and temporal dynamics must be learned primarily from limited robot interaction data~\citep{pi0-experiment-wild}.
In parallel, recent work has explored video generation models as world models for robotics~\citep{chenLargeVideoPlanner2025, paiMimicvideoVideoActionModels2025, kimCosmosPolicyFineTuning2026,yeWorldActionModels,biMotusUnifiedLatent2025}. Unlike static image-text pairs, videos provide rich spatiotemporal cues about state transitions and object interactions. This capability has catalyzed an emerging decoupled paradigm: a video world model first serves as a visual planner, generating a future visual trajectory conditioned on current observations and language instructions; subsequently, an inverse dynamics model (IDM) extracts the corresponding robot actions from the generated frames~\citep{du2023learning,bharadhwaj2024gen2act,tanAnyPosAutomatedTaskAgnostic2025}.  By separating high-level spatiotemporal reasoning from low-level control, this formulation offers a highly promising route toward scalable robot learning grounded in internet-scale video data.

Despite this promise, we identify a critical limitation in this decoupled pipeline: the absence of explicit executability constraints. We define executability as the extent to which a generated video trajectory can be translated into motor commands that accomplish the intended task while respecting the robot's physical and kinematic constraints. Foundation video generative models~\citep{sora2024,wan2025} are not explicitly trained to respect these constraints. Generated robot motion may exhibit arm deformation, self-intersections, or abrupt temporal discontinuities, which can translate into joint jumps or high-frequency jitter when decoded by an IDM. Even when such artifacts are not visually apparent, the decoded actions may still violate robot motion limits. We term this mismatch between generated robot behavior and embodiment-specific execution requirements the \textit{executability gap}. Inference-time techniques such as rejection sampling can mitigate this gap, but require generating and evaluating multiple costly video candidates.

Our key idea is to use the IDM not only to decode actions for execution, but also to provide action-space feedback for post-training the video generator. We call this framework \textbf{Executable Video Alignment (EVA)}. We train the IDM on robot demonstrations and freeze it to decode actions from generated videos. From these actions, we construct a reward that (1) penalizes non-smooth trajectories through acceleration and jerk, and (2) penalizes violations of embodiment-specific velocity and acceleration limits. Aggregating these penalties over the sequence provides a trajectory-level reward that serves as a surrogate for executability. Inspired by the recent success of reinforcement learning (RL) in foundation-model alignment~\citep{wu2025rlvr,wang2026tagrpo,cai2025phygdpo}, we use this feedback to guide updates to the video generator, incorporating action-space constraints into its rollout distribution without executing each training rollout on a robot.

We evaluate our method on RoboTwin~\citep{chen2025robotwin}, LIBERO~\citep{liuLIBERO2023}, and a real-world robotic platform. We assess both the visual quality of generated rollouts and task success when executing their IDM-decoded actions. With the same IDM before and after alignment, the post-trained video model produces rollouts with fewer embodiment-related artifacts and improves downstream execution success. We further evaluate EVA as a data engine, showing that its generated video--action pairs improve downstream VLA performance when used as training data.

Our main contributions are summarized as follows:

\begin{itemize}
    \item We identify and characterize the \textit{executability gap} in video-based robotic planners: generated robot motion and its IDM-decoded actions may fail to respect the target robot's kinematic and embodiment constraints.

    \item We propose an IDM-based executability reward for aligning video world models. A frozen IDM translates generated videos into actions, whose smoothness and violations of velocity and acceleration limits provide feedback for RL post-training of the video generator.

    \item Experiments in simulation and on a real-robot platform show improved visual quality and execution success with the same IDM, as well as more effective synthetic training data for downstream VLAs.
\end{itemize}

\vspace{5pt}

\section{Related Works}

\noindent \textbf{World Models for Policy Learning.} World models connect future visual predictions with action generation for robot policies. One paradigm separates visual prediction from action decoding: UniPi and Vidar generate video plans and use an inverse dynamics model (IDM) to recover robot actions~\citep{du2023learning,fengVidarEmbodiedVideo2025}, while VPP and mimic-video condition action generation on predictive video representations~\citep{huVideoPredictionPolicy2025,paiMimicvideoVideoActionModels2025}. Related approaches translate generated futures into control through dense correspondences or goal-conditioned policies~\citep{ko2023learning,black2023susie,bharadhwaj2024gen2act}. A second paradigm couples prediction and policy learning. VidMan transfers implicit video dynamics to action prediction, while Seer jointly learns future visual prediction and inverse dynamics~\citep{wenVidManExploitingImplicit2024,tianPredictiveInverseDynamics2024}. Unified video--action and world action models generate visual futures and actions through shared backbones or interacting experts~\citep{zhuUnifiedWorldModels2025,cenWorldVLAAutoregressiveAction2025,biMotusUnifiedLatent2025,kimCosmosPolicyFineTuning2026,yeWorldActionModels}. EVA follows the video-generation-and-IDM paradigm, extending the IDM from an action decoder to a source of feedback for video-model post-training.

\noindent \textbf{Video Generation for Robotics.} Beyond serving as policy components, video models provide synthetic experience for robot learning and evaluation. Video generators synthesize diverse scenes and interactions~\citep{sora2024,wan2025,agarwal2025cosmos}. As data engines, DreamGen and GigaWorld generate manipulation videos paired with inferred actions to expand policy-training data~\citep{jangDreamGenUnlockingGeneralization2025,teamGigaWorld0WorldModels2025}. As learned simulators, action-conditioned models predict the visual consequences of robot commands, supporting interaction modeling and policy evaluation~\citep{zhuIRASimFineGrainedWorld2025,guoCtrlWorldControllableGenerative2025,liaoGenieEnvisionerUnified2025,teamEvaluatingGeminiRobotics2025}. Both uses require generated robot motion to be consistent and controllable. CLOVER regularizes video prediction with optical flow, while RoboMaster adds motion guidance and object appearance and shape conditioning~\citep{buClosedLoopVisuomotorControl2024,fuLearningVideoGeneration2025}. EVA complements these approaches by improving both the visual integrity and action executability of synthetic training pairs.

\noindent \textbf{Alignment and Post-Training in Generative Models.} RL aligns generative models with human feedback~\citep{ouyang2022training} or automated rewards~\citep{guo2025deepseek}, commonly using PPO~\citep{schulman2017proximal} or GRPO~\citep{deepseek-math}. Flow-GRPO extends reward-based optimization to flow-matching models~\citep{liu2025flow}. For video world models, Cosmos-Predict uses VLM-based feedback on text alignment, motion quality, and visual quality~\citep{nvidiaWorldSimulationVideo2025,liu2025improving}, while ReWorld learns multi-dimensional rewards for physical realism, task completion, embodiment plausibility, and visual quality~\citep{pengReWorldMultiDimensionalReward2026}. RLIR rewards agreement between IDM-recovered actions and the actions conditioning video generation~\citep{ye2025rlir}. EVA instead targets language-conditioned video planning without target action trajectories, deriving feedback from IDM-decoded actions to encourage temporal smoothness and embodiment-specific motion constraints.

\section{Problem Setup}
\label{sec:problem_setup}

Given recent camera observations and a language instruction $c=(o_{t-h+1:t},\ell)$, a video planner samples a future rollout $V\sim p_\theta(\cdot\mid c)$. An inverse dynamics model (IDM) $f_\phi$ maps temporal windows of $V$ to a robot command sequence $A$. Our objective is to improve the video generator so that its rollouts exhibit fewer embodiment-related artifacts and support more successful task execution under the same action decoder.

Our backbone is a latent flow-matching video generator~\citep{wan2025,lipman2022flow}. A video VAE maps a training video to $z_1$, and $z_0\sim\mathcal{N}(0,\mathbf{I})$ supplies noise. With flow time $s\in[0,1]$, the interpolation is $z_s=(1-s)z_0+sz_1$, and supervised training minimizes
\begin{equation}
\mathcal{L}_{\mathrm{FM}}
=\mathbb{E}_{z_0,z_1,s,c}
\left[\left\|u_\theta(z_s,s,c)-(z_1-z_0)\right\|_2^2\right].
\label{eq:fm_setup}
\end{equation}
At inference time, ODE sampling transforms noise into a latent video, which the VAE decodes into frames.

\section{Executable Video Alignment}
\label{sec:eva}

\subsection{Inverse Dynamics Model}
\label{sec:idm}

The IDM infers robot control commands from a short temporal window of visual observations. Given frames $I_{t-k:t+k}$ centered at time $t$, the IDM predicts the joint-position command $q_t$. We train the IDM on robot trajectory data with supervised regression:
\begin{equation}
\mathcal{L}_{\mathrm{IDM}}
=\mathbb{E}\left[\sum_t\left\|f_\phi(I_{t-k:t+k})-q_t^{\mathrm{gt}}\right\|_2^2\right],
\label{eq:idm_training}
\end{equation}
where $k$ denotes the temporal context radius.

The IDM uses a convolutional backbone, spatial softmax, and an MLP action head~\citep{levine2016end}. Architecture details and the spatial-representation ablation are provided in Appendix~\ref{subsec:idm_ablation}. The trained IDM is frozen throughout video-model post-training.

\begin{figure*}[t]
    \centering
    \includegraphics[width=\linewidth]{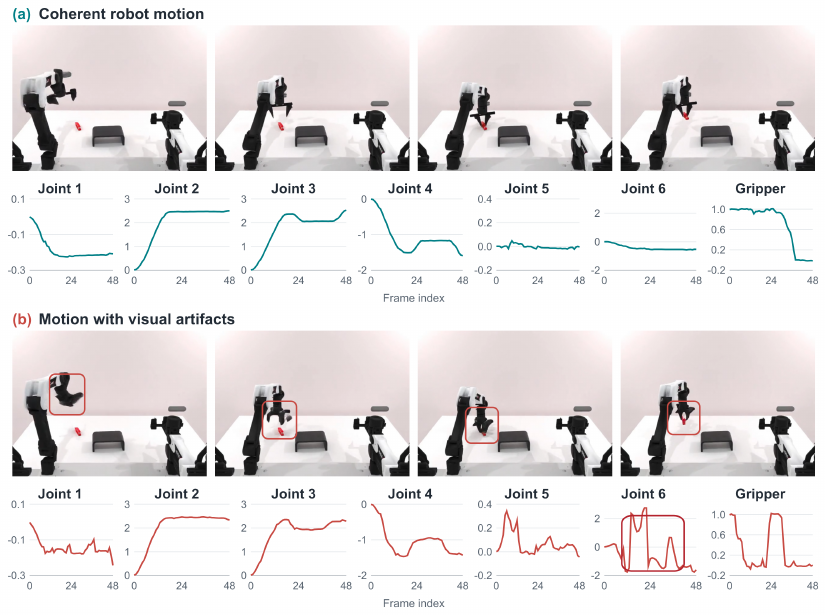}
    \caption{IDM-decoded actions from generated videos with and without visual artifacts. Plots show six left-arm joint-position commands and one gripper command, ordered from base to gripper. \textbf{Top:} coherent robot motion with smooth decoded commands. \textbf{Bottom:} visual artifacts (red boxes) accompanied by command fluctuations, especially in the distal joint and gripper (dimensions 6--7).}
    \label{fig:IDM_case}
\end{figure*}

\subsection{IDM-Based Executability Reward}
\label{sec:rl_post}

We assess executability through the smoothness and motion-limit violations of IDM-decoded actions, using them as a surrogate for downstream execution (\Cref{fig:IDM_case}).

Given a generated video $V$, the frozen IDM predicts a sequence of joint-position commands $A=\{q_t\}_{t=1}^{T}$ at control interval $\Delta t$. We compute velocity $v_t=(q_{t+1}-q_t)/\Delta t$, acceleration $\alpha_t=(v_{t+1}-v_t)/\Delta t$, and jerk $j_t=(\alpha_{t+1}-\alpha_t)/\Delta t$. To penalize non-smooth motions, we apply a robust Huber penalty:
\begin{equation}
\mathrm{Huber}(x;\delta)=
\begin{cases}
\frac{1}{2}x^2,&|x|\le\delta,\\
\delta\left(|x|-\frac{1}{2}\delta\right),&|x|>\delta,
\end{cases}
\label{eq:huber}
\end{equation}
yielding acceleration and jerk penalties
\begin{equation}
\mathcal{P}_{\alpha}=\mathbb{E}_t[\mathrm{Huber}(\alpha_t;\delta_\alpha)],\qquad
\mathcal{P}_{j}=\mathbb{E}_t[\mathrm{Huber}(j_t;\delta_j)].
\label{eq:smoothness_penalties}
\end{equation}
The Huber loss is applied componentwise and aggregated over command dimensions. We further penalize violations of the robot's velocity and acceleration bounds:
\begin{equation}
\mathcal{P}_{\mathrm{vel}}=\mathbb{E}_t\left\|\max(|v_t|-v_{\max},0)\right\|_2^2,\qquad
\mathcal{P}_{\mathrm{acc}}=\mathbb{E}_t\left\|\max(|\alpha_t|-\alpha_{\max},0)\right\|_2^2.
\label{eq:limit_penalties}
\end{equation}
The reward is the negative total penalty
\begin{equation}
R = -\left(\lambda_j\mathcal{P}_j+\lambda_\alpha\mathcal{P}_\alpha
+\lambda_{v\text{-lim}}\mathcal{P}_{\mathrm{vel}}
+\lambda_{a\text{-lim}}\mathcal{P}_{\mathrm{acc}}
\right),
\label{eq:eva_reward}
\end{equation}
where the nonnegative weights balance the components. 
Maximizing this reward favors smoother decoded motions with fewer constraint violations. The feedback is computed from generated videos without executing each training rollout on a robot.

\subsection{Post-Training and Deployment}
\label{sec:grpo_alignment}

For RL post-training with the IDM-based reward, we adopt Flow-GRPO~\citep{liu2025flow} for policy optimization. For each conditioning context, we sample $G$ videos and compute rewards $R_i$ with the frozen IDM. The group-relative advantage~\citep{deepseek-math} is $\widehat{A}_i=(R_i-\mu_R)/(\sigma_R+\epsilon)$, where $\mu_R$ and $\sigma_R$ denote the mean and standard deviation of the group rewards. The generator is then optimized using the clipped objective
\begin{equation}
\mathbb{E}\!\left[
\frac{1}{G}\sum_{i=1}^{G}
\min\!\left(r_i(\theta)\widehat{A}_i,
\mathrm{clip}(r_i(\theta),1-\varepsilon,1+\varepsilon)\widehat{A}_i\right)
-\beta D_{\mathrm{KL}}(\pi_\theta\,\|\,\pi_{\mathrm{ref}})
\right].
\label{eq:eva_grpo}
\end{equation}
Here $r_i(\theta)$ is the current-to-behavior policy transition likelihood ratio, with the sampling-step index omitted; the expectation also averages over sampled transitions and the KL term regularizes deviation from the reference policy. Only generator LoRA~\citep{hu2021lora} parameters are updated. The fine-tuned model follows the same sampling algorithm as the reference model.

At inference time, the aligned video model follows standard video-planning procedures. Longer videos are generated by conditioning each new chunk on preceding frames. For closed-loop control, the IDM decodes each chunk into robot actions; after execution, the model replans from the latest camera observation.

\begingroup
\setlength{\textfloatsep}{12pt plus 2pt minus 2pt}
\setlength{\floatsep}{10pt plus 2pt minus 2pt}
\setlength{\intextsep}{10pt plus 2pt minus 2pt}

\section{Experiments}
\suppressfloats[t]

\begin{figure}[t]
\centering
\includegraphics[width=0.98\linewidth]{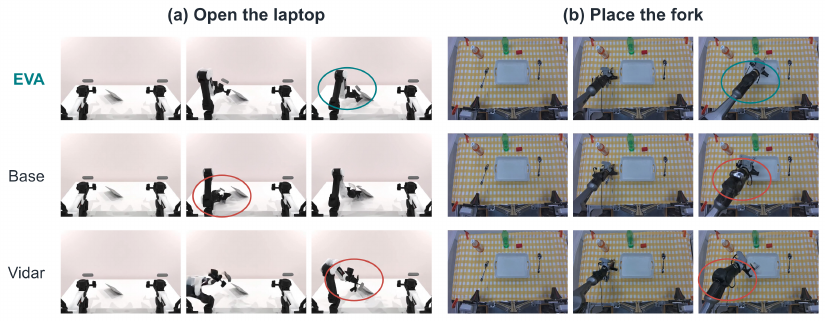}
\caption{Qualitative video comparison on two tasks. Rows show EVA, Base, and Vidar; red circles highlight arm artifacts in Base and Vidar, while EVA shows more coherent robot geometry.}
\label{fig:compare}
\vspace{-3mm}
\end{figure}

We evaluate whether action-space feedback improves three aspects of robotic video generation: visual rollout quality, task success when executing IDM-decoded actions, and the utility of generated video--action pairs for downstream VLA training.

\subsection{Experimental Settings}
\label{subsec:exp_settings}

\textbf{Models and training.} We initialize the Wan2.1-14B video backbone~\citep{wan2025} from the pretrained Large Video Planner~\citep{chenLargeVideoPlanner2025} and perform embodiment-specific supervised fine-tuning with diffusion forcing~\citep{chen2024diffusion}, yielding the base model \textbf{Base}. Starting from this checkpoint, we use EVA to perform RL post-training with the IDM-based reward, yielding the fine-tuned model \textbf{EVA}. The IDM is trained on robot demonstrations and remains frozen throughout post-training. Training configurations are provided in Appendix~\ref{sec:implementation_details}.

\begin{table}[!t]
\centering
\caption{Visual planning quality across 210 prompts, including Vidar~\citep{fengVidarEmbodiedVideo2025}. Entries are human-rated pass rates (\%); Perfect requires all criteria and visually depicted task completion.}
\label{tab:visual_quality}
\smallskip
\small
\setlength{\tabcolsep}{8pt}
\renewcommand{\arraystretch}{1.12}
\begin{tabular}{lrrrr}
\toprule
Method & Kinematic & Interaction & Instruction & Perfect \\
\midrule
Vidar (Wan2.2) & 67.6 & 66.7 & 87.6 & 62.9 \\
Base & 70.5 & 83.3 & \textbf{90.5} & 68.1 \\
EVA & \textbf{91.4} & \textbf{86.2} & 89.5 & \textbf{83.8} \\
\bottomrule
\end{tabular}
\end{table}

\textbf{Benchmarks and data.} On RoboTwin 2.0~\citep{chen2025robotwin}, we train the video model on 1,050 demonstrations across 21 tasks and evaluate Base and EVA on these tasks under the clean setting (\texttt{demo\_clean}), with 50 trials per task. On LIBERO~\citep{liuLIBERO2023}, we train both the video model and IDM on demonstrations from the Spatial, Goal, and Object suites and report execution success separately for each suite. For real-robot experiments, we use the AgileX CobotMagic platform and collect 250 training demonstrations covering five tasks. We evaluate on these five seen tasks and five additional out-of-distribution (OOD) tasks, with 20 trials per task.

\textbf{Evaluation and baselines.} Execution follows the closed-loop protocol in \Cref{sec:grpo_alignment}; success requires task completion without human intervention or safety interruption. Our main comparison is Base versus EVA. We additionally compare visual quality with Vidar (Wan2.2-5B)~\citep{fengVidarEmbodiedVideo2025} and real-robot execution with ACT, $\pi_0$, Vidar, and GE-Act (\Cref{fig:real_robot_success}). Full RoboTwin baseline comparisons and per-task results are provided in Appendix~\ref{sec:full_direct_execution}.

\subsection{Visual Rollout Quality}
\label{subsec:visual_quality}

We use structured human evaluation to assess whether action-space feedback reduces embodiment-related artifacts in generated videos. The evaluation contains 210 held-out observation--instruction prompts, with ten prompts per RoboTwin task. Generated videos are anonymized and shuffled, then rated for \emph{kinematic plausibility} (arm visual integrity and temporal continuity), \emph{interaction plausibility} (contact and object motion), and \emph{instruction following}. A rollout is rated \emph{perfect} if it satisfies all three criteria and visually depicts task completion. This rubric evaluates the generated plan; actual execution success is assessed separately below.

Kinematic plausibility improves most, with a 20.9\% gain over Base (\Cref{tab:visual_quality}), while interaction plausibility increases modestly and instruction following remains similar. This pattern is consistent with the reward's focus on decoded motion: action-space feedback can discourage visually inconsistent robot motion without directly supervising task semantics or object interactions. The improvement in visual quality suggests that the frozen IDM provides a useful connection between action constraints and video generation. Qualitative examples (Fig.~\ref{fig:compare}) show the reduction in arm deformation.

\subsection{Direct Action Execution}
\label{subsec:sim_validation}
\label{subsec:real_world}

\textbf{Simulation.} EVA improves mean success by 6.48\% on RoboTwin and by 6.5--7.5 points across the three LIBERO suites (\Cref{tab:simulation_success}). On RoboTwin, success improves on 18 of 21 tasks (Appendix~\ref{sec:full_direct_execution}). Together with the visual evaluation, these gains suggest that more coherent generated robot motion can translate into more reliable downstream execution: reducing motion discontinuities in visual plans may help the IDM decode smoother action sequences that are easier to execute. A similar pattern is observed across all three LIBERO suites, where EVA consistently improves execution success over Base.

\begin{figure}[!t]
\centering
\includegraphics[width=0.96\linewidth]{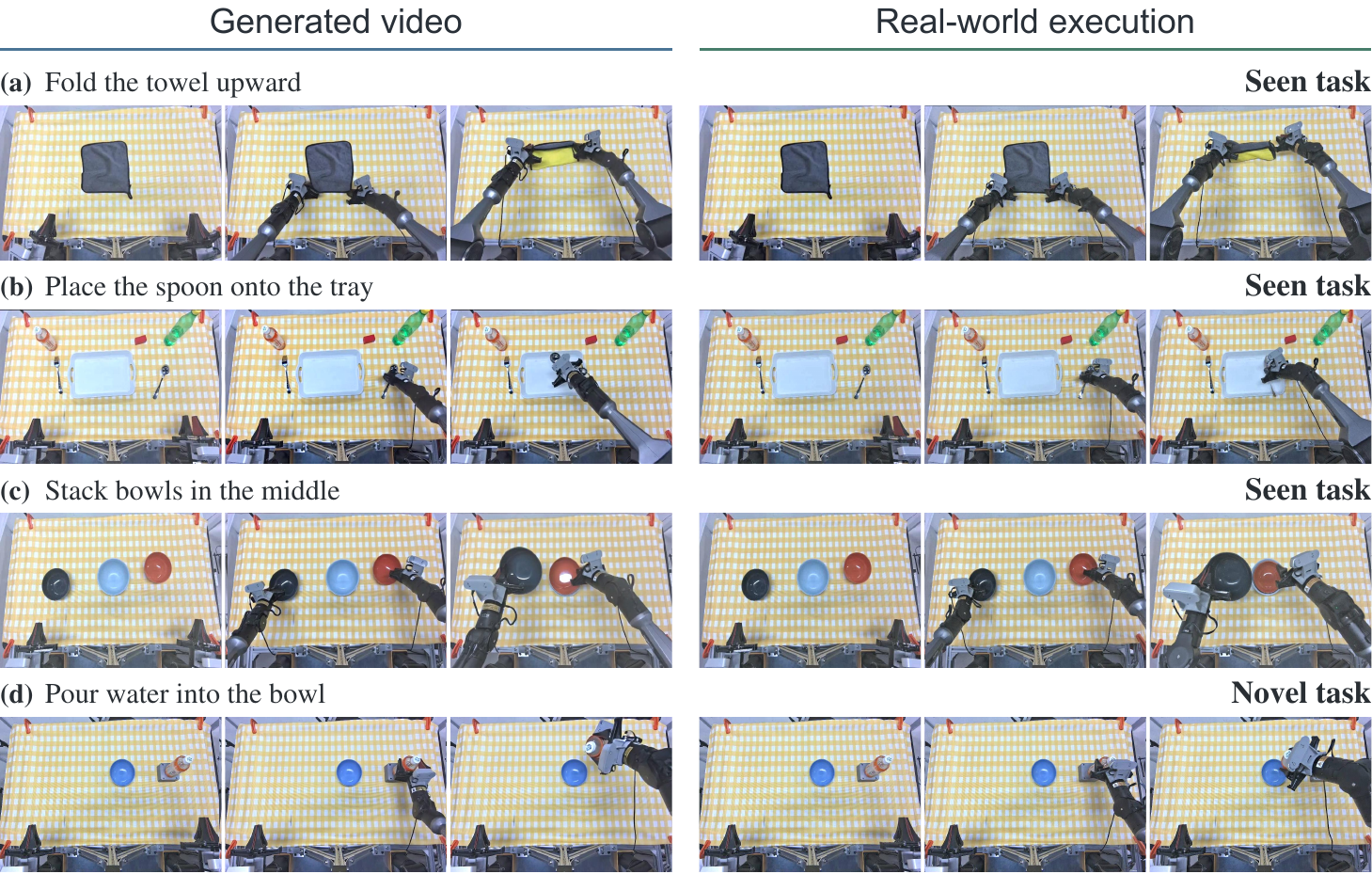}
\caption{Real-robot deployment. Left: Generated video plans ; Right: corresponding robot executions.}
\label{fig:real_world}
\end{figure}

\begin{table}[!htbp]
\centering
\caption{Direct execution success. RoboTwin reports successes/50 trials for three representative tasks and mean success (\%) over all 21 tasks under the clean setting. LIBERO reports success rates (\%) over 200 trials per suite. $\Delta$: EVA minus Base, in counts for individual RoboTwin tasks and percentage points otherwise.}
\label{tab:simulation_success}
\smallskip
\small
\setlength{\tabcolsep}{5pt}
\renewcommand{\arraystretch}{1.10}
\begin{tabular}{@{}lrrrrrrrr@{}}
\toprule
\multirow{2}{*}{Method} & \multicolumn{5}{c}{RoboTwin 2.0} & \multicolumn{3}{c}{LIBERO} \\
\cmidrule(lr){2-6}\cmidrule(l){7-9}
& \makecell{Place\\Mouse} & \makecell{Place Obj.\\Stand} & \makecell{Turn\\Switch} & $\cdots$ & \makecell{Mean (\%)\\21 tasks} & Spatial & Goal & Object \\
\midrule
Base & 10/50 & 28/50 & 23/50 & $\cdots$ & 45.81 & 46.5 & 49.5 & 63.5 \\
\textbf{EVA} & \textbf{15/50} & \textbf{33/50} & \textbf{30/50} & $\cdots$ & \textbf{52.29} & \textbf{53.5} & \textbf{57.0} & \textbf{70.0} \\
$\Delta$ & +5 & +5 & +7 & $\cdots$ & +6.48 & +7.0 & +7.5 & +6.5 \\
\bottomrule
\end{tabular}
\end{table}

\textbf{Real-robot deployment.} The video-planning pipeline achieves higher overall success than the evaluated VLA baseline $\pi_0$, with a particularly clear advantage on OOD tasks (\Cref{fig:real_robot_success}). Applying EVA post-training increases execution success on both seen and OOD tasks. Smoother commands that remain within motion limits may make generated plans easier to execute on the physical robot, reducing failures caused by jitter and abrupt joint motions that the robot cannot reliably follow. The gains on OOD tasks suggest that these motion-level benefits can extend beyond the training tasks. \Cref{fig:real_world} pairs generated plans with their physical executions.

\begin{figure}[!t]
\centering
\includegraphics[width=\linewidth]{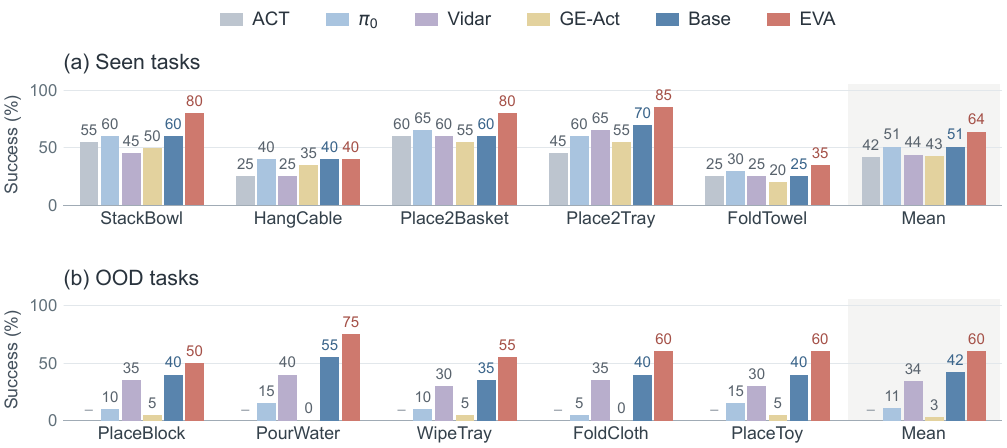}
\caption{Real-robot success (\%) on seen and OOD tasks with 20 trials each: ACT~\citep{zhao2023learning}, $\pi_0$~\citep{black2024pi_0}, Vidar~\citep{fengVidarEmbodiedVideo2025}, and GE-Act~\citep{liaoGenieEnvisionerUnified2025}. Shaded groups show means; dashes denote unreported ACT OOD results.}
\label{fig:real_robot_success}
\end{figure}

\subsection{Synthetic Data for VLA Training}
\label{sec:worldarena_data_engine}

Following WorldArena's data-engine evaluation framework~\citep{shang2026worldarena}, we fine-tune $\pi_{0.5}$~\citep{physicalintelligence2025pi05} on Base- or EVA-generated videos paired with IDM-decoded actions for Click Bell, Place A2B Right, and Place Container Plate. Both conditions use the same data quantity and policy-training settings. Each policy is evaluated over 100 trials per task (\Cref{tab:worldarena_vla}); implementation details are provided in Appendix~\ref{sec:implementation_details}.

\begin{table}[!htbp]
\centering
\caption{Downstream $\pi_{0.5}$ success after fine-tuning on Base- or EVA-generated data for three RoboTwin tasks. Entries are success rates (\%) over 100 trials per task.}
\label{tab:worldarena_vla}
\smallskip
\small
\setlength{\tabcolsep}{5pt}
\begin{tabular*}{0.90\linewidth}{@{\extracolsep{\fill}}lcccc@{}}
\toprule
Data source & \makecell{Click\\Bell} & \makecell{Place A2B\\Right} & \makecell{Place Container\\Plate} & Mean \\
\midrule
Base-generated data & 92 & 18 & 53 & 54.3 \\
EVA-generated data & \textbf{98} & \textbf{26} & \textbf{67} & \textbf{63.7} \\
\bottomrule
\end{tabular*}
\end{table}

With matched data quantity and policy-training settings, EVA-generated data improves $\pi_{0.5}$ success on all three tasks, increasing mean success from 54.3\% to 63.7\% (\Cref{tab:worldarena_vla}). These results provide initial evidence that executability alignment can improve the usefulness of world-model-generated data for downstream robot learning. Appendix~\ref{sec:supp_data_engine} further shows that EVA can generate plausible rollouts from synthetic initial scenes, although the downstream training utility of this fully synthetic setting remains to be evaluated.

\subsection{Ablations and Analysis}
\label{subsec:alignment_analysis}

Auxiliary training is evaluated on all 21 RoboTwin tasks; reward ablation and inference-time controls use separate three-task evaluations.

\textbf{Reward diagnostics.} Artifact-free rollouts and successful executions generally receive higher rewards (Appendix~\Cref{fig:reward_visual}), suggesting that decoded actions expose motion inconsistencies relevant to both generation and execution. However, nearly static motion incurs little smoothness penalty, and incorrect interactions can still yield smooth, in-limit commands (Appendix~\Cref{fig:reward_hacking}). The reward therefore guides how the robot moves without directly assessing whether it achieves the intended interaction.

\textbf{Reward components.} Smoothness penalties yield higher execution success than motion limits alone, despite similar visual perfect rates (\Cref{tab:reward_ablation_main}b). Commands can fluctuate within admissible bounds, so limit penalties alone may leave jitter that is difficult to execute. Conversely, smooth motion can still exceed robot limits. The full reward performs best, supporting the complementary roles of temporal smoothness and motion-limit constraints.

\textbf{Training and inference-time strategies.} We examine how the proposed action-space feedback can be used during training and inference (\Cref{tab:reward_ablation_main}a). IDM-aux SFT augments the supervised video objective with the negative of the same IDM-based reward used by EVA. Starting from the same Base checkpoint and using the same robot data with comparable optimization budgets, both methods improve execution success, supporting action-space feedback as a useful alignment signal across optimization objectives. EVA directly optimizes this reward on model-generated rollouts without paired target videos during post-training. It thus applies feedback to the same free-running generation process encountered at inference, while achieving modestly higher success than auxiliary SFT.

We also compare inference-time interventions. Local smoothing around motion-limit violations or abrupt command changes does not improve execution success, suggesting that such corrections cannot reliably repair motion errors in generated rollouts and may distort task-relevant movements. Reward-based rejection sampling improves Base by selecting the best of four candidates, demonstrating the reward's utility for inference-time selection. However, it requires multiple video generations and IDM evaluations per planning round. EVA incorporates the same feedback during post-training and achieves slightly higher success with a single candidate. Together, these results support IDM-derived action-space feedback across training and inference, with RL integrating this signal into free-running video generation and avoiding the additional sampling cost of inference-time selection.

\begin{table}[!htbp]
\centering
\caption{Ablations and alternative strategies on RoboTwin. (a) Auxiliary training uses 21 tasks; inference-time controls use a separate three-task evaluation. Base + Smooth smooths decoded actions; $G_{\mathrm{test}}$ denotes candidates per planning round. (b) Reward ablation uses another three-task evaluation. (c) EVA applied to Vidar~\citep{fengVidarEmbodiedVideo2025} on OpenLaptop. All rates are percentages.}
\label{tab:reward_ablation_main}
\smallskip
\hspace*{-4pt}%
\begin{minipage}[t]{0.45\linewidth}
\centering
\small
\textbf{(a) Training and inference}\par\smallskip
\setlength{\tabcolsep}{5pt}
\renewcommand{\arraystretch}{1.0836}
\begin{tabular*}{0.86\linewidth}{@{\extracolsep{\fill}}lrr@{}}
\toprule
Method & $G_{\mathrm{test}}$ & Success (\%) \\
\midrule
\multicolumn{3}{@{}l}{\textit{Auxiliary training (21 tasks)}} \\
Base & 1 & 45.81 \\
IDM-aux SFT & 1 & 50.29 \\
EVA & 1 & \textbf{52.29} \\
\midrule
\addlinespace
\multicolumn{3}{@{}l}{\textit{Inference-time controls (3 tasks)}} \\
Base & 1 & 53.0 \\
Base + Smooth & 1 & 51.7 \\
Base + RS-4 & 4 & 63.3 \\
EVA & 1 & \textbf{65.0} \\
\bottomrule
\end{tabular*}
\end{minipage}\hspace{0.06\linewidth}%
\begin{minipage}[t]{0.49\linewidth}
\centering
\small
\textbf{(b) Reward components}\par\smallskip
\setlength{\tabcolsep}{4pt}
\renewcommand{\arraystretch}{1.15}
\begin{tabular*}{0.94\linewidth}{@{\extracolsep{\fill}}lrr@{}}
\toprule
Reward setting & \makecell{Video\\perfect (\%)} & \makecell{Execution\\(\%)} \\
\midrule
Base & 63.3 & 46.67 \\
Acc.\ + jerk only & 76.7 & 63.33 \\
Limits only & 75.0 & 56.67 \\
Full reward & \textbf{81.7} & \textbf{68.33} \\
\midrule
\multicolumn{3}{c}{\textbf{(c) EVA applied to Vidar}} \\
\midrule
Task & Vidar & Vidar + EVA \\
\midrule
OpenLaptop & 50.0 & \textbf{65.0} \\
\bottomrule
\end{tabular*}
\end{minipage}\hspace*{4pt}
\end{table}

\textbf{Transfer across video models and action decoders.} Applying EVA to Vidar improves success on OpenLaptop (\Cref{tab:reward_ablation_main}c), providing initial evidence of applicability to another video model. EVA also improves success by 6.2 percentage points with an independently trained ViT-based IDM never used for reward computation, supporting transfer beyond the reward-producing decoder (Appendix~\ref{subsec:idm_ablation}).

\section{Conclusion}

We presented \textbf{Executable Video Alignment (EVA)}, a framework for addressing the executability gap in robotic video generation through action-space feedback. A frozen IDM decodes the motion implied by generated rollouts, and penalties on non-smooth motion and velocity- and acceleration-limit violations guide reinforcement-learning updates to the video generator. Experiments on RoboTwin, LIBERO, and a real bimanual robot show reduced embodiment-related visual artifacts and improved task success when executing IDM-decoded actions. EVA-generated video--action pairs also improve downstream VLA performance when used as training data. These results support action-space feedback as a practical way to improve robotic video models for both direct control and synthetic data generation.

\noindent \textbf{Limitations and Future Work.} Our reward focuses on kinematic feasibility and smoothness, and does not explicitly model contact dynamics such as forces, friction, or torques, which are critical for precision contact-rich manipulation. Diffusion-based video generation is also computationally expensive in our current setup, limiting applicability to high-frequency reactive control. Future work will explore richer dynamics-aware reward signals and faster sampling (e.g., distillation or fewer-step samplers) to enable more responsive closed-loop deployment.

\par\endgroup

\subsection*{AI use statement}
AI tools assisted with literature research for Related Work and language polishing.

\bibliographystyle{eva_reference}
\bibliography{exported_items}

\clearpage
\appendix

\renewcommand{\thefigure}{S\arabic{figure}}
\renewcommand{\thetable}{S\arabic{table}}
\renewcommand{\theHfigure}{supp.\arabic{figure}}
\renewcommand{\theHtable}{supp.\arabic{table}}
\setcounter{figure}{0}
\setcounter{table}{0}

\raggedbottom %
\section{Failure Modes and Additional Qualitative Results}
\label{sec:additional_execution}

We examine failures in rollouts generated by unaligned video models during real-robot evaluation. Visual inconsistencies can propagate through the IDM into unstable commands, while incorrect interactions or goals can prevent task completion even when the motion appears plausible.

\begin{figure}[H]
    \centering
    \includegraphics[width=1.0\linewidth]{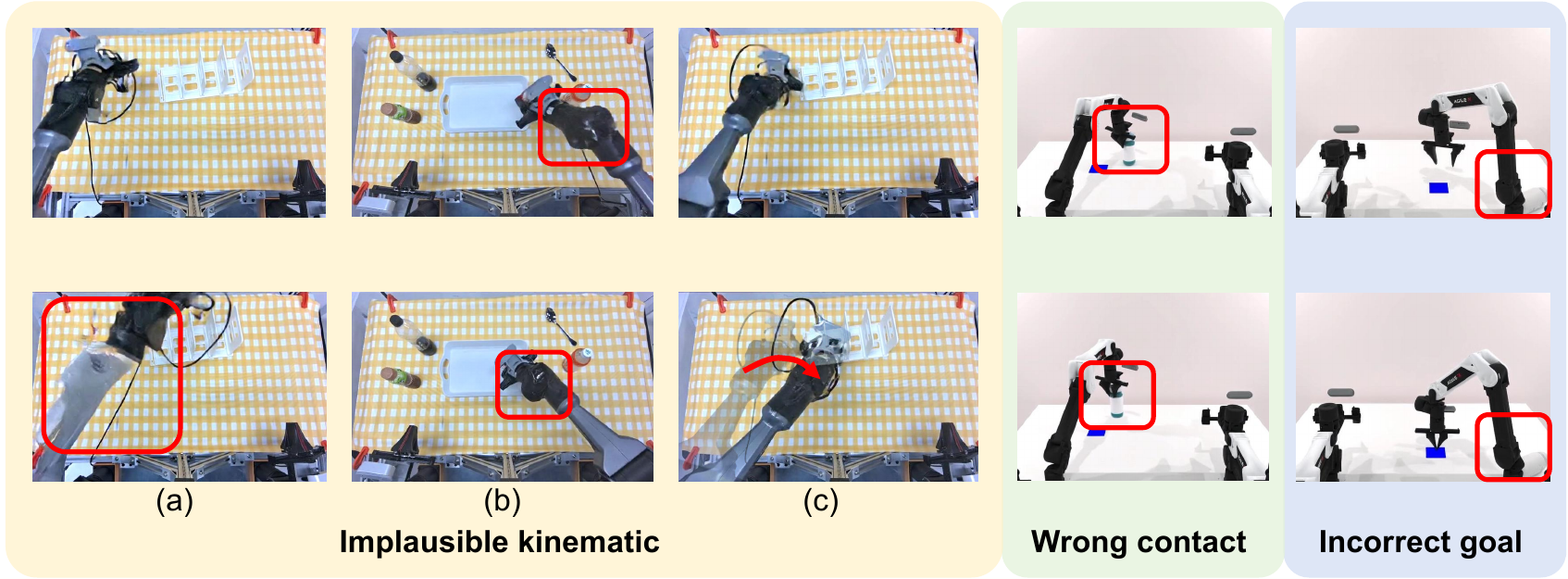}
    \caption{Common failure modes observed in unaligned video world models during real-world execution. \textbf{Implausible kinematics:} violations of rigid-body consistency, including (a) morphological deformation, (b) ambiguous joint articulation, and (c) temporal discontinuity. \textbf{Wrong contact:} physically inconsistent object interaction. \textbf{Incorrect goal:} failure to make progress toward the instruction-conditioned objective.}
    \label{fig:failure_mode}
\end{figure}

\Cref{fig:failure_mode} illustrates representative examples. We group the dominant failure patterns into three categories. \textbf{Implausible kinematics} refers to violations of rigid-body consistency, including morphological deformation of the robot arm, ambiguous joint articulation, and abrupt temporal discontinuities across frames. \textbf{Wrong contact} denotes physically inconsistent object interactions, such as penetration or missing contact events. \textbf{Incorrect goal} occurs when the generated rollout fails to make meaningful progress toward the instruction-conditioned objective.

\begin{figure}[H]
    \centering
    \includegraphics[width=\linewidth]{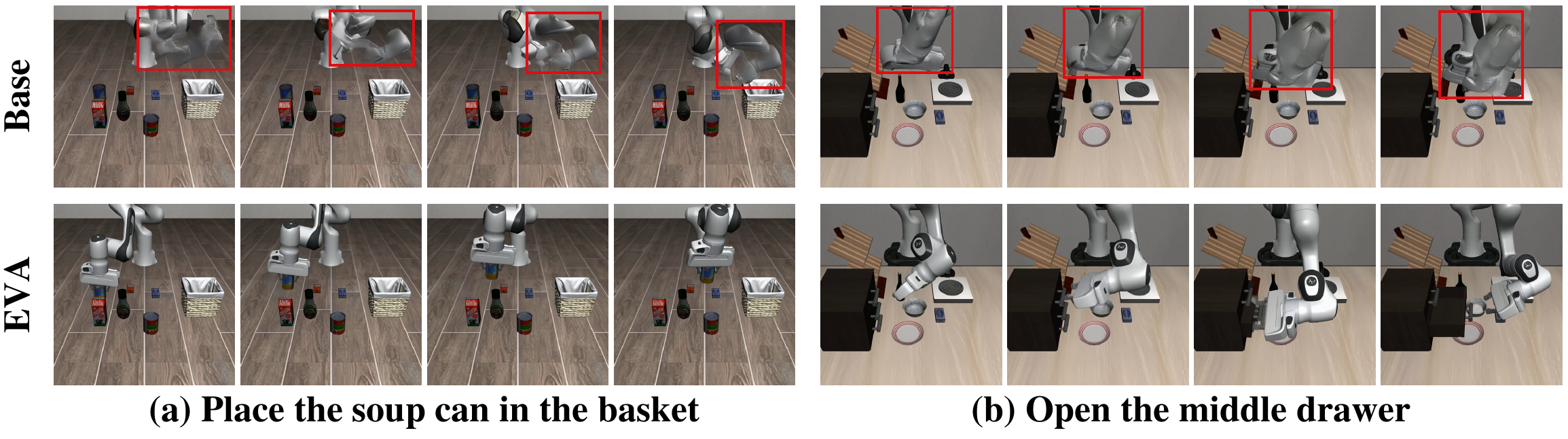}
    \caption{Generated video comparison on LIBERO. Base is shown in the top row and EVA in the bottom row. The left four columns show placing the soup can in the basket (LIBERO-Object), and the right four show opening the middle drawer (LIBERO-Goal). Panel (a) samples frames 12, 24, 36, and 48 in chronological order from the third generated chunk of each model. Panel (b) shows representative Base artifacts from an earlier chunk and EVA interaction stages from subsequent planning rounds, rather than temporally matched frames. Red boxes highlight visible arm deformation in Base rollouts.}
    \label{fig:libero_base_eva}
\end{figure}

\textbf{LIBERO visual comparison.}
\Cref{fig:libero_base_eva} compares Base and EVA rollouts on two LIBERO tasks. The soup-can example follows a single planning round for each model, while the drawer example illustrates Base artifacts and later EVA interaction stages. Red boxes identify visible arm deformation; the caption specifies the frame selection for each panel.

\section{Reward Diagnostics}
\label{sec:reward}

We examine whether the IDM-based reward distinguishes rollouts with different visual quality and execution outcomes. Both analyses use three representative tasks, with ten rollouts per group for each task.

\begin{figure}[H]
    \centering
    \includegraphics[width=1.0\linewidth]{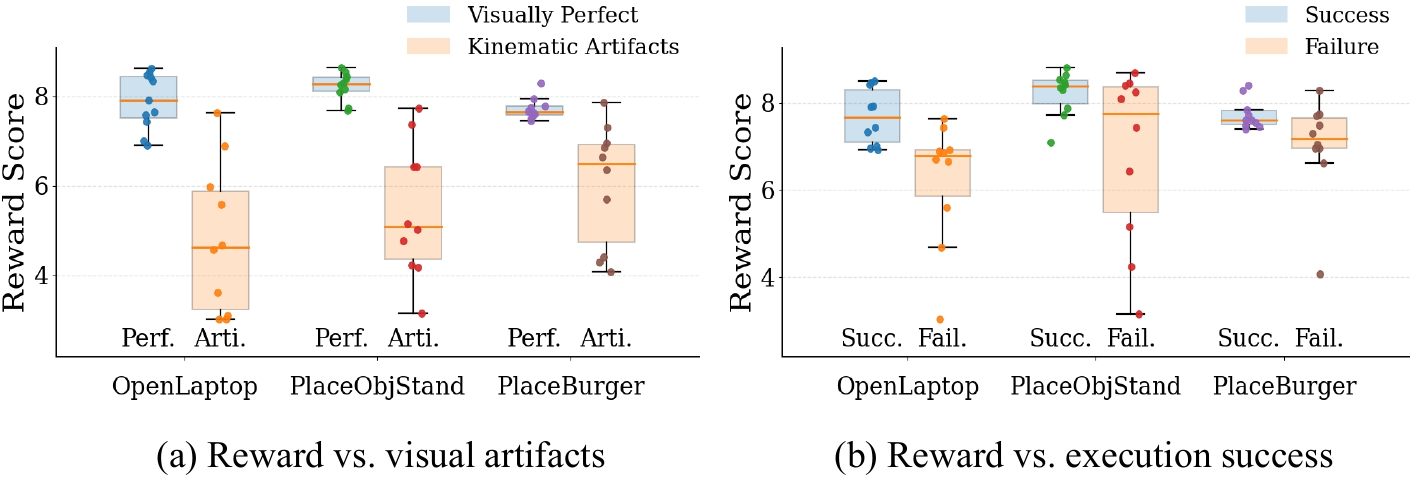}
    \caption{Relationship between reward scores and rollout quality. 
(a) Reward scores grouped by whether the generated rollout contains visible embodiment-related artifacts. 
(b) Reward scores grouped by downstream execution outcomes in simulation. Results are shown for three representative tasks, with 10 rollouts per group for each task. 
Each point represents one rollout, and box plots summarize the reward distributions.}

    \label{fig:reward_visual}
\end{figure}

\textbf{Visual quality.} We compare visually plausible rollouts with rollouts exhibiting embodiment-related kinematic artifacts. Artifact-free videos generally receive higher scores in \Cref{fig:reward_visual}(a), suggesting that the reward is sensitive to inconsistencies in generated robot motion.

\textbf{Execution outcomes.} We decode generated videos into actions and execute them in simulation, then compare ten successful and ten failed rollouts per task. Successful executions generally receive higher scores in \Cref{fig:reward_visual}(b). Together, these observations support the reward as a useful surrogate within the sampled cases; they do not establish that a high reward guarantees task completion.

\section{Implementation and Training Details}
\label{sec:implementation_details}

\textbf{Supervised fine-tuning.} We first fine-tune the video generation model using supervised learning on 8 NVIDIA A800 GPUs with Fully Sharded Data Parallel (FSDP) and mixed-precision training in bf16. Each training sample consists of a 49-frame video clip. For the RoboTwin simulation dataset, the model is trained for approximately 4,500 optimization steps at an input resolution of $640\times480$.
For the real-world dataset collected on the physical robot platform, the model is trained for approximately 2,580 optimization steps at a resolution of $832\times480$.
The per-device batch size is set to 1 with gradient accumulation of 4 steps, giving a total batch size of 32. 
We use the AdamW optimizer with learning rate $8\times10^{-6}$, $\beta_1=0.9$, $\beta_2=0.95$, and weight decay of 0.05. 
The learning rate follows a constant schedule with 100 warm-up steps. 
We additionally enable diffusion forcing with random history conditioning during training to improve temporal consistency.

\textbf{GRPO post-training.} After supervised fine-tuning, we further align the model using GRPO on eight NVIDIA A800 GPUs with a total batch size of 32.
We optimize the policy with AdamW using a learning rate of $2\times10^{-4}$, $\beta_1=0.9$, $\beta_2=0.95$, and weight decay of 0.05.
For each prompt, 8 candidate videos are sampled for policy optimization.
The clipping range is set to $0.001$, with advantage clipping at $5.0$ and a maximum gradient norm of $1.0$.
We also apply KL regularization with coefficient $\beta=0.004$ during optimization.
Only the LoRA parameters (rank 32) are updated, while the backbone remains frozen.
In the simulation setting, GRPO is trained for 136 optimization steps with 4 inner policy epochs per iteration, taking approximately 6 days.
In the real-world setting, we use 2 inner policy epochs and train for 46 optimization steps, taking approximately 3 days.
The corresponding training reward curves are shown in \Cref{fig:reward_curve}.

For RoboTwin and the real-world setting, we use $(\lambda_j,\lambda_\alpha,\lambda_{v\text{-lim}},\lambda_{a\text{-lim}})=(0.5,0.25,2.0,1.0)$ and $\Delta t=1/16\,\mathrm{s}$. For LIBERO, whose OSC controller operates at 20\,Hz, we use $\Delta t=1/20\,\mathrm{s}$. The Huber thresholds are $\delta_\alpha=0.5\alpha_{\max}$ and $\delta_j=0.5\alpha_{\max}/\Delta t$. Rewards are computed from unstandardized actions, with gripper dimensions weighted at $0.2$ relative to arm joints. The reward is the negative weighted sum of the four penalties defined in the main text.

\begingroup
\setlength{\intextsep}{3pt}
\begin{figure}[H]
    \centering
    \includegraphics[width=0.88\linewidth]{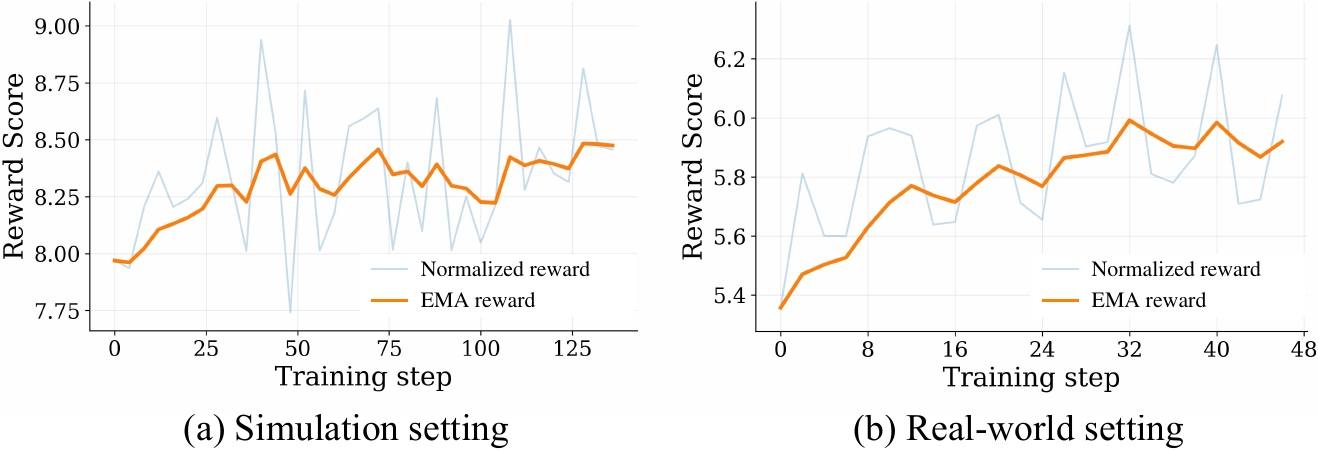}
    \caption{Training reward curves in simulation and real-robot experiments. Rewards are normalized to $[0,10]$ for visualization. The blue curve shows the normalized reward, while the orange curve shows its exponential moving average (EMA). 
    In both settings, the reward exhibits a generally increasing trend during alignment, showing the evolution of the optimization objective.}
    \label{fig:reward_curve}
\end{figure}

\begin{figure}[H]
\noindent\begin{minipage}[t]{0.48\linewidth}
    \vspace{0pt}
\textbf{LIBERO implementation details.} The IDM predicts a 7-D command $q_t=(p_t,r_t,g_t)$, comprising absolute end-effector position $p_t\in\mathbb{R}^3$, absolute rotation-vector orientation $r_t\in\mathbb{R}^3$, and gripper command $g_t$. At 20\,Hz, each target pose is converted into an incremental operational-space control (OSC) command relative to the observed pose, using the shortest relative rotation. Rewards are computed before OSC conversion from component-wise first-, second-, and third-order differences of $q_t$ with $\Delta t=1/20\,\mathrm{s}$, following Sec.~4.2. The soft coordinate-wise thresholds are
\[
\begin{aligned}
v_{\max}&=[0.75,0.75,0.75,4,4,4,30],\\
a_{\max}&=[10,10,10,50,50,50,450].
\end{aligned}
\]
For position, rotation-vector, and gripper channels, first-order units are $\mathrm{m/s}$, $\mathrm{rad/s}$, and $\mathrm{s}^{-1}$; second-order units are $\mathrm{m/s^2}$, $\mathrm{rad/s^2}$, and $\mathrm{s}^{-2}$. Rotational differences serve as coordinate-space smoothness proxies. The gripper weight is $0.2$ relative to pose channels. Each suite is evaluated over 200 trials.

\end{minipage}\hfill
\begin{minipage}[t]{0.48\linewidth}
    \vspace{0pt}
    \centering
    \includegraphics[width=0.90\linewidth]{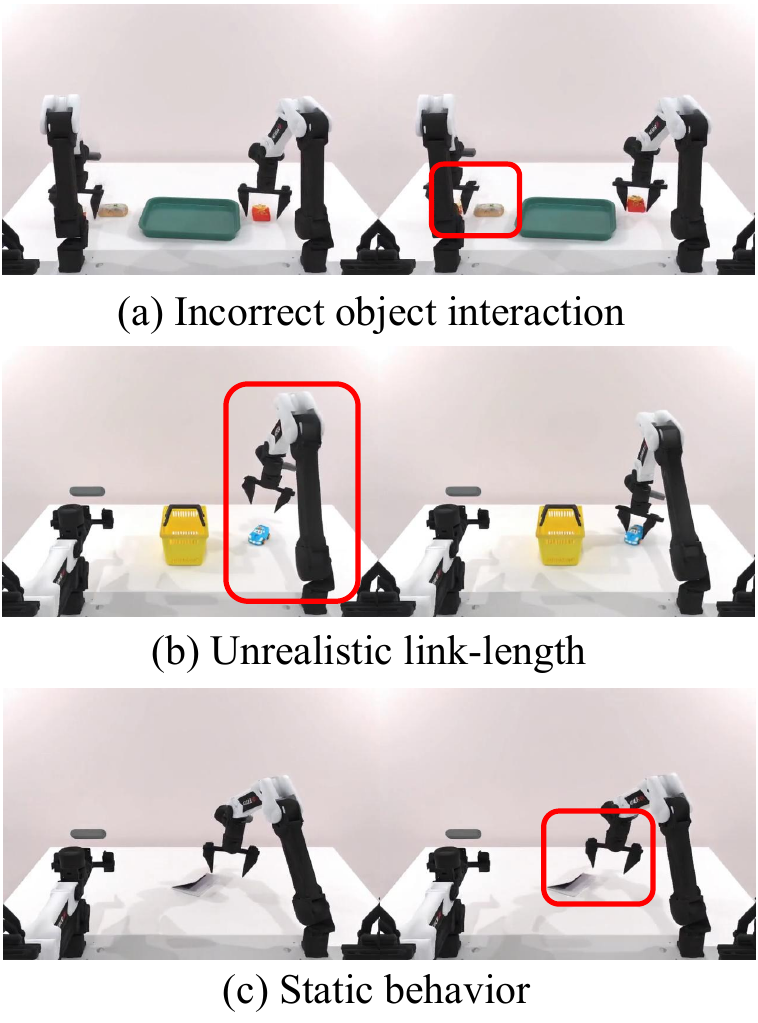}
    \caption{High-reward failure cases during GRPO post-training.}
    \label{fig:reward_hacking}
\end{minipage}
\end{figure}
\endgroup

\textbf{Inference settings.} Long-horizon video generation uses 49-frame chunks and conditions the next chunk on the last four frames of the preceding chunk. In closed-loop control, the IDM translates each generated chunk into an action chunk, which is executed before the next chunk is generated from the resulting camera observation. These two modes share the same generator and IDM but differ in whether the generated chunk is only concatenated into a video or executed on the robot.

\textbf{Reward hacking.} The IDM-based reward is designed as a surrogate for motion-level executability rather than task completion. During the early and intermediate stages of post-training, increasing reward is generally accompanied by improvements in robot-motion integrity and downstream execution. With prolonged optimization, however, the proxy can be over-optimized: we observe high-reward failures including incorrect contacts, unrealistic robot geometry, and nearly static rollouts that do not complete the task (\Cref{fig:reward_hacking}).

To mitigate this behavior, we use KL regularization to constrain deviation from the reference video model and monitor a held-out validation set that is excluded from both policy optimization and final test evaluation. During training, we qualitatively assess two validation criteria: \textbf{robot integrity}, which captures embodiment consistency and temporal continuity, and \textbf{task progress}, which assesses whether generated rollouts make meaningful progress toward the instructed goal. We stop optimization once improvements on these validation criteria have largely saturated, before continued reward maximization leads to persistent visible degeneration. The checkpoint at this stopping point is used for evaluation; no downstream test success is used for stopping or model selection.

\textbf{Synthetic data and VLA training.} For each data source in \Cref{tab:worldarena_vla}, Base and EVA, we generate 50 trajectories per task for Click Bell, Place A2B Right, and Place Container Plate, yielding 150 trajectories per source. Generation starts from the head-camera images of the corresponding demonstration episodes, and each generated video matches its source episode's frame count. The frozen IDM decodes the generated videos into actions. For each data source, we initialize a separate $\pi_{0.5}$ policy from the same official pretrained checkpoint and jointly fine-tune it on all three tasks. Both conditions use 10,000 training steps, a batch size of 64, an action horizon of 40, and eight GPUs with FSDP. The policy uses only head-camera images, with wrist-camera inputs disabled. Evaluation uses the RoboTwin clean setting with the same head-camera configuration as the source data, with 100 trials per task.

\section{IDM Architecture and Ablation}
\label{subsec:idm_ablation}

\paragraph{Architecture.} The IDM follows a standard visuomotor design~\citep{levine2016end}: an ImageNet-pretrained ResNet-50~\citep{he2016resnet} extracts spatial features, a spatial softmax layer converts each channel into a 2D coordinate, and an MLP maps these coordinates to actions. To accommodate temporally stacked RGB frames, we expand the first convolution along the input-channel dimension, initializing its weights by repeating the pretrained filters and dividing by the number of input frames. The MLP has two hidden layers of widths 1,024 and 512. Let $F\in\mathbb{R}^{C\times H\times W}$ denote the feature map obtained from temporally stacked input frames. For feature channel $m$, spatial softmax is defined as
\begin{equation}
p_{ij}^{m}=\frac{\exp(F_{ij}^{m})}{\sum_{i',j'}\exp(F_{i'j'}^{m})},\qquad
(x_m,y_m)=\sum_{i,j}p_{ij}^{m}(i,j).
\label{eq:spatial_softmax}
\end{equation}
The coordinates $\{(x_m,y_m)\}_{m=1}^{C}$ are concatenated and fed into an MLP to predict $q_t$.

\paragraph{Spatial-representation ablation.}

We evaluate the IDM's spatial representation by replacing spatial softmax with global average pooling. Test accuracy is the fraction of predicted action vectors for which all 14 dimensions satisfy the error thresholds simultaneously: absolute error below $0.05$\,rad for each arm joint and below $0.5$ for each $0/1$-encoded gripper command. We also evaluate task success by executing the decoded actions in simulation.

As shown in \Cref{tab:idm_ablation}, replacing spatial softmax reduces prediction accuracy from 0.9864 to 0.7738 and execution success from 89.52\% to 84.29\%. This comparison supports retaining explicit spatial features in the action decoder.

\begin{table}[H]
\centering
\caption{Ablation study on the IDM architecture. Test accuracy requires all 14 action dimensions to have absolute errors below $0.05$\,rad for arm joints and $0.5$ for $0/1$ gripper commands simultaneously. Execution success is measured in simulation.}
\vspace{3pt}
\label{tab:idm_ablation}
\setlength{\tabcolsep}{5pt}
\renewcommand{\arraystretch}{1.2}
\begin{tabular}{lcc}
\toprule
\textbf{Inverse Dynamics Model} & \textbf{Test Accuracy} & \textbf{Test Success Rate} \\
\midrule
Spatial softmax & 0.9864 & 89.52\% \\
Global average pooling & 0.7738 & 84.29\% \\
\bottomrule
\end{tabular}
\end{table}

\paragraph{Cross-architecture IDM evaluation.}
Since the same IDM provides rewards and decodes actions in our main experiments, we test whether EVA's gains persist with a decoder that did not guide alignment. The evaluation IDM uses a ViT-S/16 backbone, spatial softmax, and an MLP action head to predict $q_t$. It is independently trained on the same robot demonstrations with the same regression objective. This IDM is used exclusively for evaluation, with no involvement in EVA post-training or reward computation. Decoding Base and EVA videos with this same ViT-IDM yields success rates of 43.5\% and 49.7\%, respectively (\Cref{tab:cross_arch_idm}). The 6.2-point gain closely matches the original IDM's 6.5-point gain, suggesting that EVA's improvements to generated motion transfer to a decoder with a different visual backbone, beyond the one supplying the reward.

\begin{table}[H]
\centering
\caption{Cross-architecture IDM evaluation. Entries report execution success rates (\%); $\Delta$ denotes the improvement in percentage points. ResNet IDM-A is used for reward computation, while ViT IDM-B is used only for evaluation.}
\label{tab:cross_arch_idm}
\vspace{3pt}
\setlength{\tabcolsep}{12pt}
\renewcommand{\arraystretch}{1.15}
\begin{tabular}{lcc}
\toprule
Video planner & ResNet IDM-A & ViT IDM-B \\
\midrule
Base & 45.8 & 43.5 \\
EVA & 52.3 & 49.7 \\
$\Delta$ & +6.5 & \textbf{+6.2} \\
\bottomrule
\end{tabular}
\end{table}

\section{RoboTwin Direct-Execution Results}
\label{sec:full_direct_execution}

\Cref{tab:full_success} compares performance on 21 RoboTwin tasks under \texttt{demo\_clean}. The baselines use single-task policies, while Base and EVA use a single multi-task video model. Relative to Base, EVA improves success on 18 tasks and matches it on two; Stamp Seal decreases from 13/50 to 8/50.

\begin{table}[H]
\centering
\caption{RoboTwin clean-setting results on 21 tasks, comparing ACT~\citep{zhao2023learning}, DP~\citep{chi2023diffusionpolicy}, RDT~\citep{liu2024rdt}, and $\pi_0$~\citep{black2024pi_0}. Baseline entries are rounded /50 equivalents of published success rates; Base and EVA report observed successes out of 50 trials. Baseline averages are computed from the rounded entries.}
\label{tab:full_success}
\vspace{3pt}
\small
\renewcommand{\arraystretch}{1.1}
\begin{tabularx}{\textwidth}{lXXXXXX}
\toprule
Task & ACT & DP & RDT & $\pi_0$ & Base & EVA \\
\midrule
Click Alarmclock & 16/50 & 31/50 & 31/50 & 32/50 & 48/50 & 50/50 \\
Click Bell & 29/50 & 27/50 & 40/50 & 22/50 & 42/50 & 48/50 \\
Handover Block & 21/50 & 5/50 & 23/50 & 23/50 & 0/50 & 0/50 \\
Handover Mic & 43/50 & 27/50 & 45/50 & 49/50 & 1/50 & 8/50 \\
Move Pillbottle Pad & 0/50 & 1/50 & 4/50 & 11/50 & 8/50 & 16/50 \\
Move Stapler Pad & 0/50 & 1/50 & 1/50 & 0/50 & 0/50 & 1/50 \\
Open Laptop & 28/50 & 25/50 & 30/50 & 43/50 & 15/50 & 25/50 \\
Place A2B Right & 0/50 & 7/50 & 1/50 & 14/50 & 26/50 & 28/50 \\
Place Bread Basket & 3/50 & 7/50 & 5/50 & 9/50 & 36/50 & 38/50 \\
Place Burger Fries & 25/50 & 36/50 & 25/50 & 40/50 & 30/50 & 32/50 \\
Place Cans Plasticbox & 8/50 & 20/50 & 3/50 & 17/50 & 20/50 & 22/50 \\
Place Container Plate & 36/50 & 21/50 & 39/50 & 44/50 & 34/50 & 40/50 \\
Place Dual Shoes & 5/50 & 4/50 & 2/50 & 8/50 & 2/50 & 3/50 \\
Place Mouse Pad & 0/50 & 0/50 & 1/50 & 4/50 & 10/50 & 15/50 \\
Place Object Basket & 8/50 & 8/50 & 17/50 & 8/50 & 6/50 & 6/50 \\
Place Object Stand & 1/50 & 11/50 & 8/50 & 18/50 & 28/50 & 33/50 \\
Press Stapler & 16/50 & 3/50 & 21/50 & 31/50 & 44/50 & 49/50 \\
Shake Bottle & 37/50 & 33/50 & 37/50 & 49/50 & 46/50 & 47/50 \\
Shake Bottle Horizontally & 32/50 & 30/50 & 42/50 & 50/50 & 49/50 & 50/50 \\
Stamp Seal & 1/50 & 1/50 & 1/50 & 2/50 & 13/50 & 8/50 \\
Turn Switch & 3/50 & 18/50 & 18/50 & 14/50 & 23/50 & 30/50 \\
\midrule
Success Rate & 29.71\% & 30.10\% & 37.52\% & 46.48\% & 45.81\% & \textbf{52.29\%} \\
\bottomrule
\end{tabularx}
\end{table}

\clearpage
\section{Real-World Experimental Setup}
\label{sec:Exp_setup}

\subsection{Robot Platform Setup}
\label{sec:supp_hardware}

Our real-world experiments are conducted on an AgileX CobotMagic platform, as shown in \Cref{fig:piper_setup} and \Cref{tab:piper_specs}. 
We use its dual-arm tabletop configuration, where each task may involve either single-arm or dual-arm manipulation depending on the task requirement.

\begingroup
\setlength{\intextsep}{6pt}
\begin{figure}[H]
\noindent\begin{minipage}[t]{0.44\linewidth}
\vspace{0pt}
\centering
\includegraphics[width=\linewidth]{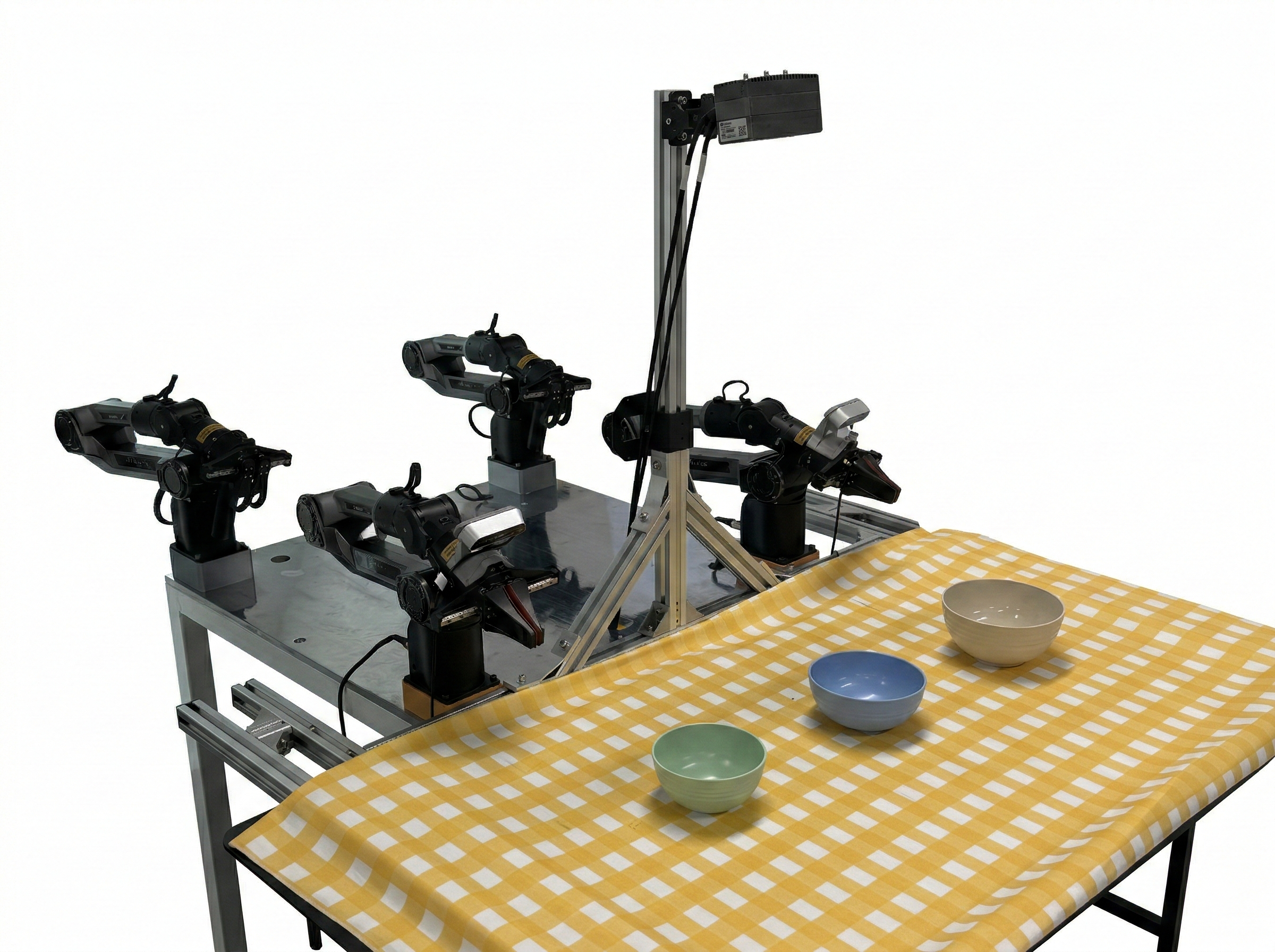}
\small
\caption{Real-world tabletop manipulation platform used in our experiments, built with AgileX PiPER robotic arms in a fixed multi-arm configuration.}
\label{fig:piper_setup}
\end{minipage}\hfill
\begin{minipage}[t]{0.53\linewidth}
\vspace{0pt}
\small
\captionof{table}{Key manipulator and gripper specifications of the AgileX PiPER arms used in our real-world experiments.}
\label{tab:piper_specs}
\smallskip
\footnotesize
\setlength{\tabcolsep}{3pt}
\renewcommand{\arraystretch}{1.0}
\begin{tabularx}{\linewidth}{@{}>{\raggedright\arraybackslash}p{0.40\linewidth}>{\raggedright\arraybackslash}X@{}}
\toprule
\textbf{Component} & \textbf{Specification} \\
\midrule
Manipulator type & AgileX PiPER \\
Arm DoF & 6 DoF per arm \\
Reach & 626.75 mm \\
Repeatability & $\pm 0.1$ mm \\
Rated payload & 1.5 kg per arm \\
Joint motion range & J1: $\pm154^\circ$; J2: $0^\circ$--$195^\circ$; J3: $-175^\circ$--$0^\circ$; J4: $-106^\circ$--$106^\circ$; J5: $-75^\circ$--$75^\circ$; J6: $\pm100^\circ$ \\
Joint max speed & J1: $180^\circ$/s; J2: $195^\circ$/s; J3: $180^\circ$/s; J4/J5/J6: $225^\circ$/s \\
Gripper type & Two-finger gripper \\
Gripper opening range & 0--70 mm \\
Gripper accuracy & $\pm 0.5$ mm \\
Rated clamping force & 40 N \\
Max clamping force & 50 N \\
\bottomrule
\end{tabularx}
\end{minipage}
\end{figure}

\subsection{Task Descriptions}
\label{sec:task}
We summarize the real-world evaluation tasks in \Cref{tab:real_world_tasks}. 
The first five tasks are seen during training; the remaining five evaluate out-of-distribution (OOD) manipulation. Each task is evaluated in 20 trials.

\newcolumntype{L}[1]{>{\raggedright\arraybackslash}m{#1}}
\newcolumntype{C}[1]{>{\centering\arraybackslash}m{#1}}
\begin{table}[H]
\centering
\caption{Real-world evaluation tasks used in our experiments.}
\vspace{3pt}
\label{tab:real_world_tasks}
\footnotesize
\setlength{\tabcolsep}{4pt}
\renewcommand{\arraystretch}{1.0}

\begin{tabular}
{@{}m{0.16\linewidth} C{0.14\linewidth} L{0.65\linewidth}@{}}
\toprule
\rule{0pt}{2.4ex}\raisebox{0.15ex}{\makecell[c]{\textbf{Task Name}}} &
\rule{0pt}{2.4ex}\raisebox{0.15ex}{\makecell[c]{\textbf{Arm Type}}} &
\rule{0pt}{2.4ex}\raisebox{0.15ex}{\makecell[c]{\textbf{Task Description}}} \\
\midrule

StackBowl
& Dual-Arm
& The robot uses both arms to grasp two bowls from the two sides and stack them onto the target bowl placed at the center of the table. \\

HangCable
& Single-Arm
& The robot uses a single arm to grasp a black cable and hang it onto the designated rack. \\

Place2Basket
& Single-Arm
& The robot uses a single arm to pick up an object from the tabletop and place it into a nearby basket. \\

Place2Tray
& Single-Arm
& The robot uses a single arm to grasp an object from the table and place it onto the tray. \\

FoldTowel
& Dual-Arm
& The robot uses both arms to grasp the two bottom corners of a towel and fold it upward. \\
\midrule

\makecell[l]{PlaceBlock\\(OOD)}
& Single-Arm
& The robot uses a single arm to pick up a block and place it into the bowl with the specified color and position. \\

\makecell[l]{PourWater\\(OOD)}
& Single-Arm
& The robot uses a single arm to grasp a bottle and pour water into a bowl on the table. \\

\makecell[l]{WipeTray\\(OOD)}
& Single-Arm
& The robot uses a single arm to grasp a towel and wipe the white tray on the tabletop. \\

\makecell[l]{FoldCloth\\(OOD)}
& Dual-Arm
& The robot uses both arms to grasp the two bottom corners of a piece of clothing and fold it upward. \\

\makecell[l]{PlaceToy\\(OOD)}
& Single-Arm
& The robot uses a single arm to grasp a soft plush toy and place it into a basket. \\
\bottomrule
\end{tabular}
\end{table}
\endgroup

\clearpage
\section{Additional Out-of-Distribution Video Generation}
\label{sec:ood_generation}

\begin{figure}[H]
    \centering
    \includegraphics[width=1.0\linewidth]{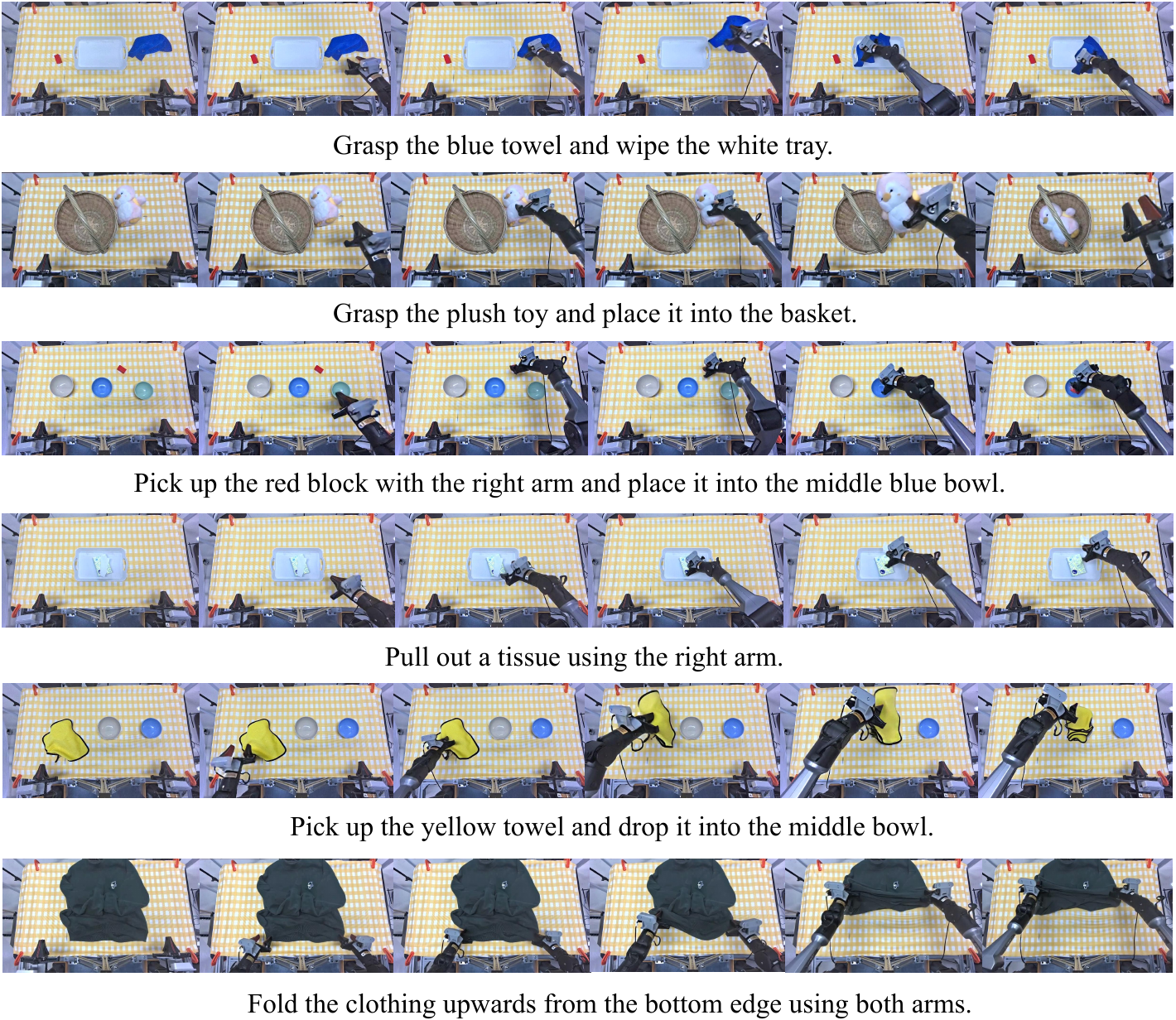}
    \caption{EVA video generation on out-of-distribution tasks. Each row shows a synthesized video sequence of six evenly sampled frames, where the first frame serves as the input conditioning image and the instructions below correspond to the task prompts.}
    \label{fig:unseen_task_vid_gen}
\end{figure}

\Cref{fig:unseen_task_vid_gen} illustrates EVA rollouts on OOD task instructions. Each sequence starts from a conditioning image and shows the generated progression. These examples assess qualitative video generation rather than measured robot execution success.

\clearpage
\section{Scaling Embodied Data via Zero-Shot Generation}
\label{sec:supp_data_engine}

\begin{figure}[H]
    \centering
    \includegraphics[width=1.0\linewidth]{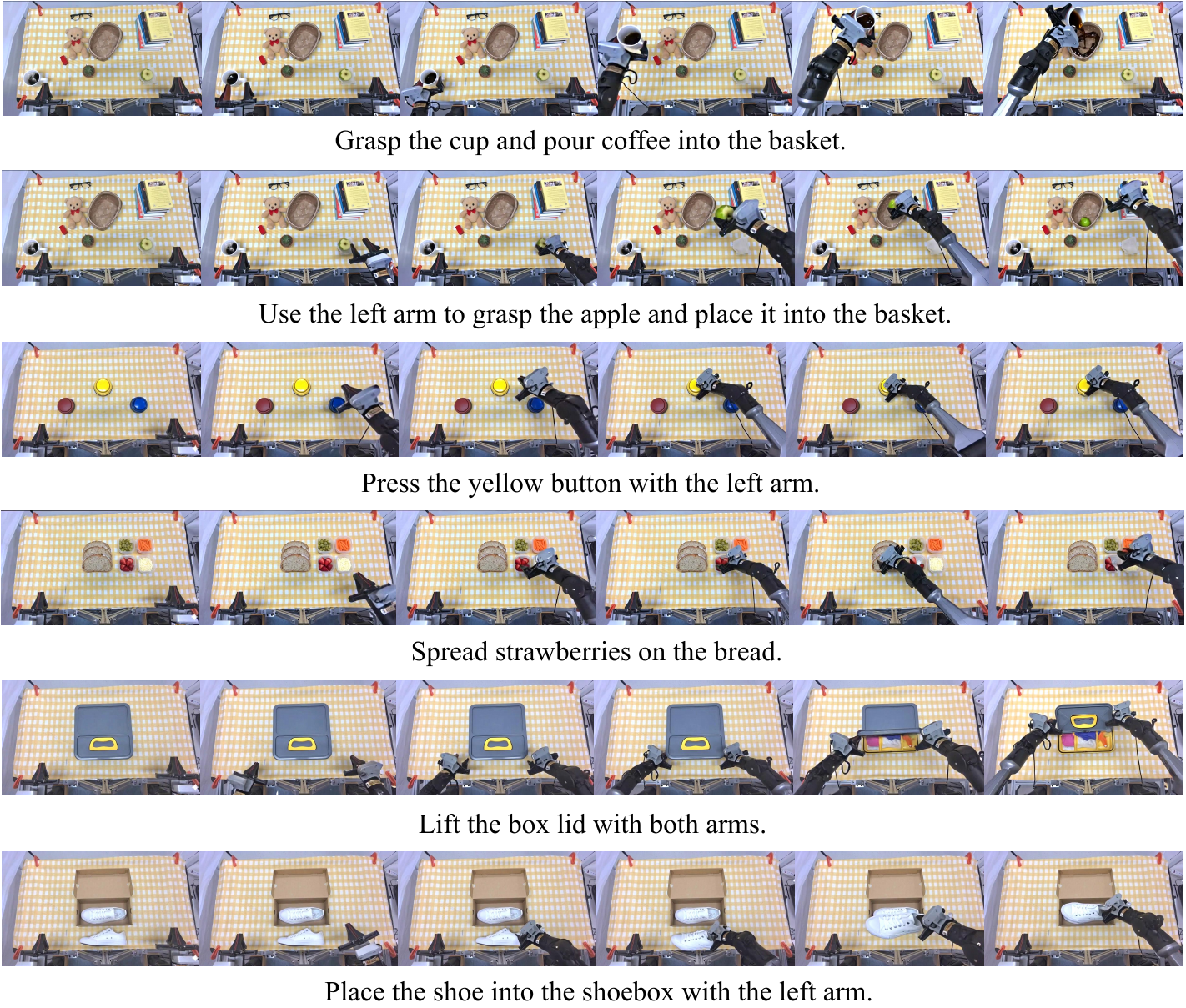}
    \caption{EVA rollouts conditioned on synthetic initial scenes. Each row shows a generated video sequence whose first frame is supplied by the text-to-image generator.}
    \label{fig:nano}
\end{figure}

We also explore a fully synthetic input pipeline for generating manipulation videos. A text-to-image generator (e.g., Nano Banana Pro~\citep{raisinghani2025nanobananapro}) produces initial scene images, which condition EVA to generate subsequent robot motion. \Cref{fig:nano} shows representative sequences.

This pipeline offers a potential source of diverse visual trajectories without collecting a teleoperated demonstration for every scene. The examples illustrate qualitative generation from synthetic observations; downstream data utility requires policy-training evaluation, as considered in the WorldArena experiment.

\end{document}